\documentclass[10pt,twocolumn,letterpaper]{article}

\usepackage{cvpr}              

\definecolor{cvprblue}{rgb}{0.21,0.49,0.74}
\usepackage[pagebackref,breaklinks,colorlinks,allcolors=cvprblue]{hyperref}
\usepackage{algorithm}
\usepackage{algpseudocode}

\def\paperID{8} 
\def\confName{CVPR}
\def\confYear{2025}

\title{GraphPad: Inference-Time 3D Scene Graph Updates for Embodied Question Answering}

\author{
Muhammad Qasim Ali\textsuperscript{*}\\
University of Waterloo\\
{\tt\small m45ali@uwaterloo.ca}
\and
Saeejith Nair\textsuperscript{*}\\
University of Waterloo\\
{\tt\small smnair@uwaterloo.ca}
\and
Alexander Wong\\
University of Waterloo\\
{\tt\small a28wong@uwaterloo.ca}
\and
Yuchen Cui\\
University of California, Los Angeles\\
{\tt\small yuchencui@cs.ucla.edu}
\and
Yuhao Chen\\
University of Waterloo\\
{\tt\small yuhao.chen1@uwaterloo.ca}
}

\begin{document}
\maketitle

\begin{abstract}

Structured scene representations are a core component of embodied agents, helping to consolidate raw sensory streams into readable, modular, and searchable formats. Due to their high computational overhead, many approaches build such representations in advance of the task.  However, when the task specifications change, such static approaches become inadequate as they may miss key objects, spatial relations, and details. We introduce \textbf{GraphPad}, a modifiable structured memory that an agent can tailor to the needs of the task through API calls. It comprises a mutable scene graph representing the environment, a navigation log indexing frame-by-frame content, and a scratchpad for task-specific notes. Together, GraphPad serves as a dynamic workspace that remains complete, current, and aligned with the agent’s immediate understanding of the scene and its task. On the OpenEQA benchmark, GraphPad attains \textbf{55.3\,\%} accuracy—\textbf{+3.0 pp} over an image‑only baseline using the same vision–language model—while operating with \textbf{five times fewer} input frames. These results show that allowing online, language‑driven refinement of 3‑D memory yields more informative representations without extra training or data collection. 

\end{abstract}

\section{Introduction}
A household robot has just scanned the kitchen when a user asks, \emph{``Is the red mug back inside the upper cupboard?''}  
Because the scene graph was built from only a handful of earlier keyframes, it lacks both a node for the mug and the relation linking it to the cupboard interior.  
With this critical information missing, the agent must either guess, rescan the entire scene, or refuse the query altogether.  
This scenario highlights a fundamental limitation in current embodied AI: \textbf{structured 3D memories are typically finalized before the downstream task is known, frequently omitting objects and spatial relations that later prove essential for action or reasoning}.

\paragraph{Inherent limitations of static memories.}
Modern embodied systems commonly pair a vision-language model (VLM) with pre-computed structured scene representations such as 3D scene graphs ~\cite{hughes2022hydra,li2024llm}.  These structured scene representations help to consolidate large amounts of dense, noisy RGB-D frames into a shared scene representation. By modeling the scene as distinct entities and relations, Scene Graphs help VLMs structure their complex multi-step spatial reasoning while performing navigation and manipulation tasks \cite{li2024llm, chiang2024mobilityvlamultimodalinstruction}.  

A core challenge with such memory systems is that crucial compression decisions—specifically, which objects to retain and what aspects of the scene to describe—are made once. When the subsequent task shifts, these pre-configured, static representations can become insufficient, incomplete, or inaccurate for the new task at hand.

This fundamental flaw is vividly demonstrated in Embodied Question Answering (EQA) \cite{Das_2018_CVPR}. Here, generic scene graphs, built without any foresight into the specific questions that will be asked, consistently perform poorly when attempting to provide answers \cite{majumdar2024openeqa}. This underscores a critical point: a memory system optimized for one general purpose simply can't adapt effectively to the nuanced demands of a different, more specific task.

Current solutions follow two primary approaches.  
\emph{Over-provisioning} retains nearly every candidate detection, incurring substantial memory and computational overhead~\cite{anwar2024remembr}.  
\emph{Offline enhancement pipelines} append heuristically chosen affordances or relations for anticipated tasks~\cite{li2024llm}, but each new domain demands additional engineering effort and cannot recover information never initially detected.  
Neither strategy scales effectively when user requests vary widely.

\paragraph{Our approach: language-guided memory updates.}
We propose that an embodied agent should be able to \emph{dynamically revise} its structured scene memory when reasoning reveals a gap in its knowledge.  
Rather than discarding an inadequate structured representation—and thereby losing its benefits—or recreating it from scratch at high computational cost, the agent should be able to modify and enhance the existing representation. This allows the agent to incorporate task-relevant information without starting over. Moreover, the agent should be able to do this through targeted perceptual queries on specific keyframes, and seamlessly integrate the newly discovered information into its scene graph. 

\paragraph{GraphPad.}
We introduce \textbf{GraphPad}, a modifiable 3D scene graph memory whose content can be updated at inference time through language-level commands issued by the same VLM that performs high-level reasoning.  
Whenever uncertainty arises or missing information is identified, the VLM can:  
(i)~retrieve additional keyframes from its navigation log,  
(ii)~insert previously undetected objects or spatial relationships, and  
(iii)~annotate existing nodes with task-specific attributes.  
These operations are exposed as language-callable functions that the reasoning agent itself can invoke, eliminating the need for task-specific reasoning code.\footnote{Generic vision modules—object detector, mask extractor, depth back-projection—remain necessary but are \emph{not} specialized for any particular downstream task.}  
Through successive targeted updates, the graph evolves from a coarse initial representation into a task-conditioned model containing precisely the information needed for the current query or action planning.

\paragraph{Contributions.}
\begin{itemize}\itemsep 0pt
    \item We formulate \emph{language-driven online editing} of structured 3D memories as a solution to the task--memory mismatch inherent in fixed scene graphs.
    \item We present \emph{GraphPad}, a system in which a single VLM both identifies knowledge gaps and updates its 3D representation via language function calls during inference.
    \item On the OpenEQA benchmark~\cite{majumdar2024openeqa}, we demonstrate that GraphPad improves spatial question answering from 52.3\% to 55.3\% while reducing the number of processed frames from 25 to 5, outperforming both image-only and static-graph baselines without additional training.
\end{itemize}

By recasting perception as an interactive, language-mediated process rather than a static preprocessing step, GraphPad enables more efficient and effective reasoning about complex 3D environments—a critical capability for agents that must bridge language understanding with physical action.

\begin{figure*}
    \centering
    \includegraphics[width=1\linewidth]{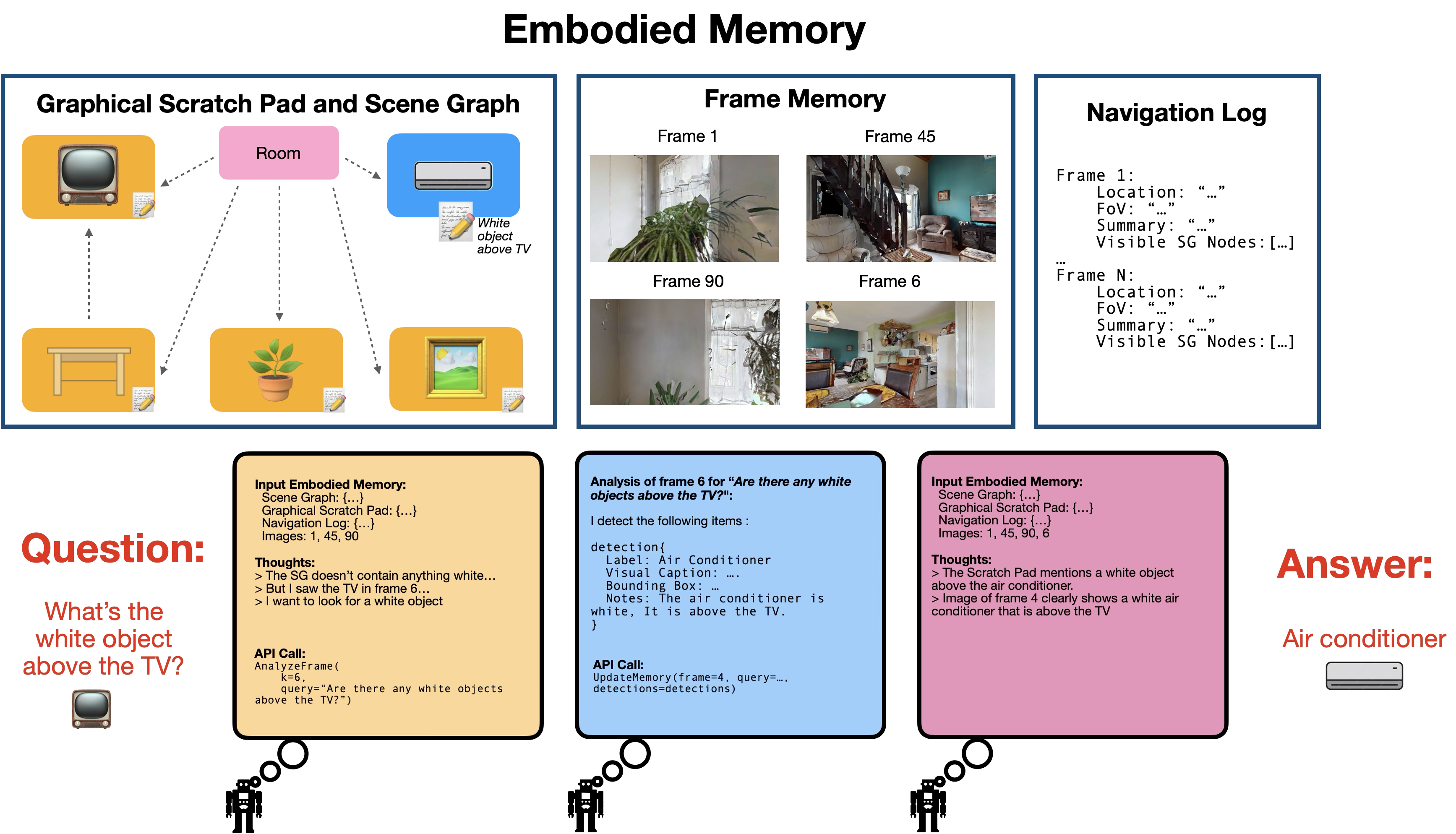}
    \caption{Overview of GraphPad for embodied question answering. The top row illustrates the Structured Scene Memory (SSM) components: Scene Graph with associated Graphical Scratch Pad, Frame Memory containing sparse keyframes, and Navigation Log indexing frame metadata. The bottom row demonstrates the inference process: given the question about a white object above the TV, the VLM first examines the initial memory state (left), identifies missing information and calls the \texttt{analyze\_frame} API on a promising frame (middle), then integrates the newly detected air conditioner into the scene graph and scratch pad before providing the answer (right). This process enables dynamic, task-specific memory updates without exhaustive preprocessing.}
    \label{fig:overview}
\end{figure*}

\section{Background and Related Work}
\label{sec:related}

Current embodied AI systems struggle with a fundamental limitation: their memory representations often lack critical information needed during task execution. This section examines key research areas relevant to this challenge.

\subsection{Embodied Memory Representations}

\textbf{Frame-level memories} store raw RGB-D streams with learned descriptors. Systems like ReMEmbR~\cite{anwar2024remembr} and Embodied-RAG~\cite{xie2024embodied} index thousands of images, allowing retrieval of relevant observations during inference. While flexible, these approaches face computational overhead from processing large image collections and delegate all spatial reasoning to language models.

\textbf{Semantic metric maps} address these limitations by embedding visual features within geometric reconstructions. VL-Maps~\cite{huang23vlmaps}, PLA~\cite{ding2023pla}, and OpenScene~\cite{peng2023openscene} support spatial grounding but lack explicit object boundaries and relationships essential for symbolic reasoning.

\textbf{3D scene graphs} compress perception into discrete entities and relations. Hydra~\cite{hughes2022hydra} pioneered real-time scene graph construction, with extensions for monocular inputs~\cite{kimera}, multi-robot fusion~\cite{greve2023curb}, and open-vocabulary labeling~\cite{gu2024conceptgraphs}. These structures align with structured VLM prompting but remain largely static after construction.

\subsection{Memory Update Mechanisms}

Several existing methodologies employ updatable memories; however, they typically fall short by not dynamically refining scene graphs with new semantic information crucial for the task at hand but absent in the initial recording:

\textbf{Dynamic object tracking} systems like DynaMem~\cite{liu2024dynamem}, OpenIN~\cite{tang2025openinopenvocabularyinstanceorientednavigation}, and Kimera~\cite{kimera} update spatial relations when objects move but rely on predefined update policies rather than reasoning-triggered modifications.

\textbf{Incremental discovery} methods like Moma-LLM~\cite{honerkamp2024language} and Search3D~\cite{takmaz2024search3d} can integrate new objects but employ fixed detection heuristics rather than responding to specific reasoning gaps.

\textbf{Memory selection} approaches like 3D-Mem~\cite{yang20243dmem3dscenememory} and KARMA~\cite{wang2024karma} retain informative views but cannot add structure absent from initial observations. Similarly, expanded-context approaches~\cite{chiang2024mobilityvlamultimodalinstruction} can reference more images but cannot insert missing relations.

\subsection{Task-Memory Alignment}

Ensuring memory representations contain task-relevant information remains a central challenge:

\textbf{Graph adaptation} techniques attempt to tailor existing memories to specific tasks. Information-theoretic approaches~\cite{maggio2024clio} compress graphs while preserving task-relevant nodes. SayPlan~\cite{rana2023sayplan} and SayNav~\cite{rajvanshi2024saynav} dynamically contract graphs during planning. Neither approach handles adding new information. LLM-enhanced scene graphs~\cite{li2024llm} augments an existing scene graph with task-specific affordances for the household rearrangement task. However, they rely on a complex scene graph enhancement pipeline that is domain-specific and not adaptable to diverse, unforeseen downstream tasks.

\textbf{Image-level reasoning} systems like Bumble~\cite{shah2024bumbleunifyingreasoningacting} and TagMap~\cite{zhang2024tag} provide direct visual access but sacrifice explicit relational structure. Graph-based reasoning methods~\cite{xie2024embodied} provide structured context but treat graphs as read-only, requiring new exploration rather than memory editing.

Empirical analyses~\cite{werby2024hierarchical} confirm that missing nodes and relations—not reasoning failures—frequently cause performance errors in embodied tasks.

\subsection{Embodied Question Answering}

Embodied Question Answering (EQA) benchmarks offer a standardized evaluation of memory adequacy. The episodic memory variant (EM-EQA)~\cite{majumdar2024openeqa} provides agents with fixed observations, isolating memory representation quality from exploration.

Recent approaches to EM-EQA highlight representation completeness as a critical factor. GraphEQA~\cite{saxena2024grapheqa} uses semantic graphs for viewpoint selection, while 3D-Mem~\cite{yang20243dmem3dscenememory} curates informative snapshots. Both systems nevertheless fail when queried entities are absent from their representations.

GraphPad addresses the task-memory alignment problem through targeted scene graph updates during inference. Unlike previous work, our approach enables the reasoning VLM to detect knowledge gaps and modify its own memory representation through three specific operations: frame retrieval, entity/relation insertion, and semantic annotation. This maintains the efficiency advantages of structured representations while addressing their typically static nature.

\section{Methodology}
\label{sec:methodology}

GraphPad builds a \textit{Structured Scene Memory} (SSM) from a sparse set of RGB-D keyframes and gives the vision-language model (VLM) that answers questions three callable functions to \emph{edit} that memory during inference. We first describe the \textbf{agentic reasoning loop} (Sec.~\ref{sec:agent}), then the four data structures that constitute the SSM (Sec.~\ref{sec:scene_memory}), and finally the \textbf{Modifiability APIs} (Sec.~\ref{sec:modifiability_apis}).

Throughout we follow the episodic-memory EQA protocol \cite{majumdar2024openeqa}: the agent is given every $k$-th RGB-D frame of a pre-recorded scan together with camera poses; no new sensing is possible after deployment.

\subsection{Agentic Reasoning Loop}
\label{sec:agent}

At test time, the VLM receives both the initial SSM and a natural language query $q$. The reasoning process unfolds iteratively: the VLM analyzes what information it needs to answer the question and repeatedly invokes Modifiability APIs to augment its understanding of the scene.

In each iteration, the VLM examines the current scene memory, identifies knowledge gaps relevant to the question, and selects: (1) which API to call, (2) which frame to analyze, and (3) what specific information to seek (expressed as a natural language query). The selected API returns new objects, relations, or semantic annotations that are integrated into the scene memory. This process continues until the VLM determines it has sufficient information or reaches a maximum of $m$ allowed API calls.

To ensure that the responses are grounded in all memory components, we prompt the VLM that it must support its final answer with dual evidence—visual evidence from frames in the Frame Memory and semantic evidence from notes in the Scratch-Pad. This constraint ensures that the VLM adequately utilizes its scene memory and searches the scene thoroughly prior to responding.

Unlike static memory systems that rely on pre-built representations, GraphPad enables a VLM to dynamically update its memory based on task requirements and its understanding of the scene. In our experiments, most questions are answered with just 2-3 targeted API calls, demonstrating the utility and efficiency of the reasoning loop.

\subsection{Structured Scene Memory}
\label{sec:scene_memory}

The SSM contains four mutually linked structures:

\begin{description}[leftmargin=0.5cm,labelindent=0cm,style=nextline]
\item[Scene Graph] A directed multigraph $G=(\mathcal{N},\mathcal{E})$ whose nodes represent object \textit{tracks}. Each node $n_i$ stores a point cloud $P_i$, pooled visual embedding $V_i$, pooled language embedding $L_i$, caption $C_i$, room/floor ID, and a list of keyframes in which the object is visible. Edges encode four view-invariant spatial relations critical for planning and manipulation: \textit{on top of}, \textit{subpart of}, \textit{contained in}, and \textit{attached to}.

\item[Graphical Scratch-Pad] Mirrors $\mathcal{N}$ but adds a free-form \texttt{notes} field initialized empty; the APIs write task-specific information here during reasoning.

\item[Frame Memory] An initial set of $n_{\mathrm{img}}$ evenly spaced keyframes. Additional frames requested by the APIs are appended (no eviction is used in our experiments).

\item[Navigation Log] For each keyframe: room, textual field-of-view tag, egocentric motion label (from pose deltas), and the set of visible node IDs. The log serves as a structured index, guiding the VLM in selecting candidate frames likely to contain information about specific objects or spatial relationships.
\end{description}

\paragraph{Initial construction.}
For every $k$-th RGB image $\mathbf{I}_t$, we run a VLM detector that outputs bounding boxes and captions. Each box is passed to SAM to obtain a mask; the mask is back-projected with depth into a point cloud and voxel-downsampled to 0.02 m. Noise is removed by keeping only the largest DBSCAN cluster (default \texttt{sklearn} parameters). Visual embeddings are extracted using CLIP ViT-L/14; language embeddings come from BGE~\cite{xiao2024cpackpackedresourcesgeneral}.

\textbf{Track association.} A new detection $D_i$ is matched to an existing track $T_j$ when the vote
\begin{equation}
S_{ij}= \mathbf{1}[V_i\cdot V_j>0.7] + \mathbf{1}[L_i\cdot L_j>0.8] + \mathbf{1}[G_{ij}>0.4]
\end{equation}
exceeds~2, where $G_{ij}$ is the fraction of points in $D_i$ within $\delta_g$ of $T_j$ (we use $\delta_g=5$ cm). This voting scheme requires at least two of the following conditions to hold with sufficient confidence: visual feature similarity, caption semantic similarity, or spatial overlap. Visual and language embeddings of the matched track are updated by an exponential moving average with $\alpha=0.5$; unmatched detections start new tracks.

\textbf{Edge discovery.} Every three frames we prompt the VLM with the current frame plus the JSON list of visible $\langle$\textit{bbox}, caption$\rangle$ pairs; the model predicts all pairwise relations among the four relation types. Each predicted edge is stored with a subject-ID, object-ID, relation label, and the VLM's free-form justification string.

\textbf{Caption consolidation.} Accumulated captions on a track are periodically compressed by prompting the VLM with the list and asking for a single sentence that faithfully paraphrases all entries.

\textbf{Room/floor labels.} We adopt the HOV-SG pipeline \cite{werby2024hierarchical}: floors from height-histogram modes and rooms via watershed segmentation on wall skeletons; room labels derive from CLIP similarity to a fixed set of class names.

When the SSM is passed to the VLM for reasoning, the Scene Graph, Scratch-Pad, and Navigation Log are serialized to JSON, while Frame Memory is supplied as interleaved images with their Frame IDs.

\subsection{Modifiability APIs}
\label{sec:modifiability_apis}

All three APIs receive a \textit{Frame ID} and a natural-language \textit{query}. Each returns a JSON patch containing new nodes, edges, or scratch-pad notes plus evidence pointers that are incorporated into the SSM.

\begin{description}[leftmargin=0.5cm,labelindent=0cm,style=nextline]
\item[\texttt{find\_objects}] Detects previously unseen instances relevant to \emph{query} in the specified frame and fuses them into $G$. The function leverages the VLM's bounding box detection capabilities to identify query-relevant objects and generate corresponding query-relevant notes. These detections are incorporated into the scene graph by associating them with existing tracks or creating new ones. The query-relevant notes are used to update the corresponding scratch-pad entries.

\item[\texttt{analyze\_objects}] For each node in the user-supplied list that is visible in the frame, the VLM analyzes its appearance and answers \emph{query}, storing the result in the node's \texttt{notes}. The function employs the VLM to examine each visible node's bounding box and generate descriptive notes pertaining to the query. These notes are stored in the corresponding nodes' scratch-pad entries. If specified nodes are not visible in the frame, the function defaults to \texttt{find\_objects}.

\item[\texttt{analyze\_frame}] A frame-level variant that jointly discovers undetected objects and annotates existing ones with respect to \emph{query}. This consolidated approach can both identify new perceptual elements and enrich the semantic understanding of known objects in a single operation.
\end{description}

These APIs enable a critical capability: the reasoning agent itself can identify and remedy gaps in its scene representation during inference. Rather than preprocessing exhaustively for anticipated questions, GraphPad builds a minimal initial representation and lets task requirements guide targeted perceptual refinement. This approach aligns with real-world robotic scenarios where complete scene understanding is computationally intractable, but targeted perception can efficiently support specific goals.

By invoking these functions, the agent patches omissions in its memory in real time, yielding a task-conditioned graph that improves both answer accuracy and confidence without requiring rescanning of the physical scene.

\section{Results}
\label{sec:results}

We evaluate GraphPad on the OpenEQA benchmark~\cite{majumdar2024openeqa} to assess how language-guided scene graph updates affect spatial reasoning performance. Our experiments use Gemini 2.0 Flash~\cite{pichai2024gemini} as both the reasoning agent and detector. All evaluations use the episodic-memory variant of OpenEQA, where systems receive fixed keyframes from 3D scene scans (HM3D~\cite{ramakrishnan2021hm3d} and ScanNet~\cite{dai2017scannet}).

\subsection{Baseline Comparison on OpenEQA}

Table~\ref{tab:openeqa_all} presents GraphPad's performance against established baselines. Using an initial frame memory size of $n_{img}=5$, search depth of $m=20$, and Frame-Level API, GraphPad achieves 55.3\% accuracy. This represents a 3.0 percentage point increase over using the same VLM (Gemini 2.0 Flash) with image-only input (52.3\%). GraphPad processes 5 initial frames compared to the 25 frames (every $k$-th frame with $k=5$ for HM3D and $k=20$ for ScanNet) used in the image-only baseline.

\begin{table}[t]
\centering
\caption{OpenEQA performance comparison across methods}
\label{tab:openeqa_all}
\begin{tabular}{r l c}
\toprule
\# & Method & Accuracy (\%) \\
\midrule
\multicolumn{3}{l}{\textbf{Blind LLMs}}\\
1 & GPT-4                           & 33.5 \\
\midrule
\multicolumn{3}{l}{\textbf{Socratic LLMs w/ Frame Captions}}\\
2 & GPT-4 w/ LLaVA-1.5              & 43.6 \\
\midrule
\multicolumn{3}{l}{\textbf{Socratic LLMs w/ Scene-Graph Captions}}\\
3 & GPT-4 w/ CG                     & 36.5 \\
4 & GPT-4 w/ SVM                    & 38.9 \\
\midrule
\multicolumn{3}{l}{\textbf{Multi-Frame VLMs}}\\
5 & GPT-4V                          & 55.3 \\
6 & Gemini-2.0 Flash                & 52.3 \\
7 & 3D-Mem (w/ GPT4V)               & \textbf{57.2} \\
\midrule
\multicolumn{3}{l}{\textbf{Human Agent}}\\
8 & Human                           & 86.8 \\
\midrule
\multicolumn{3}{l}{\textbf{Our Results}}\\
9 & GraphPad (w/ Gemini 2.0 Flash)  & 55.3 \\
\bottomrule
\end{tabular}
\vspace{-0.3cm}
\end{table}

GraphPad scores higher than static scene graph methods (CG: 36.5\%, SVM: 38.9\%), suggesting potential benefits of dynamic scene graph updates during inference. The system performs identically to GPT-4V (55.3\%) despite using a different base VLM. 3D-Mem~\cite{yang20243dmem3dscenememory} achieves 57.2\% accuracy, outperforming our approach by 1.9 percentage points, though it uses GPT-4V rather than Gemini 2.0.

The comparison suggests that structured memory with targeted updates can help bridge the performance gap between different vision-language models while potentially reducing the number of frames that need to be processed.

\subsection{Component Analysis}

To understand the contribution of different GraphPad components, we conducted an ablation study (Table~\ref{tab:ablation}) using a subset of 184 OpenEQA questions. Starting with just four raw frames (32.9\% accuracy), we progressively added components to measure their individual impact.

\begin{table}[t]
\centering
\footnotesize
\caption{Ablation study of system components}
\label{tab:ablation}
\begin{tabular}{p{5cm} c}
\toprule
Method & Accuracy (\%) \\
\midrule
Frame Memory                                          & 32.9 \\
Frame Memory + Scene Graph                            & 34.6 \\
Frame Memory + Navigation Log                         & 42.5 \\
Frame Memory + SG + Navigation Log                    & 46.9 \\
Frame Memory + Navigation Log + Image API             & 45.1 \\
Frame Memory + SG + Navigation Log + Node-level API   & 47.1 \\
Frame Memory + SG + Navigation Log + Frame-level API  & \textbf{50.5} \\
\bottomrule
\end{tabular}
\vspace{-0.3cm}
\end{table}

The Navigation Log provides the largest individual improvement (+9.6 percentage points over frame-only), suggesting that structured information about frame contents helps guide the VLM's attention. Adding a static scene graph yields a modest improvement (+1.7 percentage points over frame-only), while the Modifiability APIs add another 3.6 percentage points over the static scene representation.

Frame-level APIs (50.5\%) outperformed node-level variants (47.1\%). We observed that when using node-level APIs, the VLM tended to rely more on \texttt{find\_objects} than \texttt{analyze\_objects}, often using the former in ways similar to \texttt{analyze\_frame}.

\subsection{Effect of Search Depth and Frame Count}

We examined how search depth ($m$) affects accuracy using the OpenEQA184 subset with $n_{img}=4$ initial frames and the Frame-Level API. Table~\ref{tab:search_depth_ablation} shows that accuracy generally increases with search depth, reaching 51.2\% at $m=20$. Interestingly, limited search ($m=1$ or $m=2$) sometimes performs worse than no search ($m=0$), possibly because preliminary updates without follow-up refinement can introduce misleading information.

\begin{table}[t]
\centering
\caption{Performance as a function of search depth}
\label{tab:search_depth_ablation}
\begin{tabular}{c c}
\toprule
Search Depth & Accuracy (\%) \\
\midrule
0   & 46.9 \\
1   & 45.8 \\
2   & 46.3 \\
3   & 48.1 \\
4   & 45.9 \\
5   & 49.7 \\
20  & \textbf{51.2} \\
\bottomrule
\end{tabular}
\vspace{-0.3cm}
\end{table}

The initial number of frames in Frame Memory also influences performance (Table~\ref{tab:visual_memory_size_ablation}). Accuracy improves as more frames are included, though gains diminish beyond 5 frames. We observed that with fewer initial frames, the model made more API calls on average, potentially compensating for the limited initial context.

\begin{table}[t]
\centering
\caption{Performance as a function of initial frame count}
\label{tab:visual_memory_size_ablation}
\begin{tabular}{c c}
\toprule
Initial Frames & Accuracy (\%) \\
\midrule
2 & 48.2 \\
3 & 47.4 \\
4 & 49.9 \\
5 & 50.3 \\
6 & \textbf{50.5} \\
\bottomrule
\end{tabular}
\vspace{-0.3cm}
\end{table}

\subsection{Performance by Question Category}

Breaking down performance by question category (Table~\ref{tab:method_vs_gemini}) reveals variation in how GraphPad (with $k=5$) compares to using Gemini-2.0 with only images (every 5th frame in the scene).

\begin{table}[t]
\centering
\caption{Performance by question category: GraphPad vs. Gemini 2.0 with images only}
\label{tab:method_vs_gemini}
\begin{tabular}{p{3.5cm} c c c}
\toprule
Category & GraphPad (\%) & Gemini (\%) \\
\midrule
Attribute Recognition      & \textbf{66.8} & 46.5 \\
Object State Recognition   & \textbf{69.6} & 66.5 \\
Functional Reasoning       & \textbf{59.2} & 53.5 \\
World Knowledge            & \textbf{55.9} & 52.0 \\
Object Recognition         & 58.4 & \textbf{62.2} \\
Spatial Understanding      & 47.7 & \textbf{52.4} \\
Object Localization        & 31.3 & \textbf{34.3} \\
\bottomrule
\end{tabular}
\vspace{-0.3cm}
\end{table}

GraphPad shows larger improvements in attribute recognition (+20.3 percentage points), functional reasoning (+5.7 percentage points), and object state recognition (+3.1 percentage points). These categories often require detailed analysis of specific object properties. The image-only baseline performs better in object recognition (-3.8 percentage points), spatial understanding (-4.7 percentage points), and object localization (-3.0 percentage points), which may rely more on raw visual processing capabilities.

\subsection{API Call Distribution}

\begin{figure}[t]
    \centering
    \includegraphics[width=0.8\linewidth]{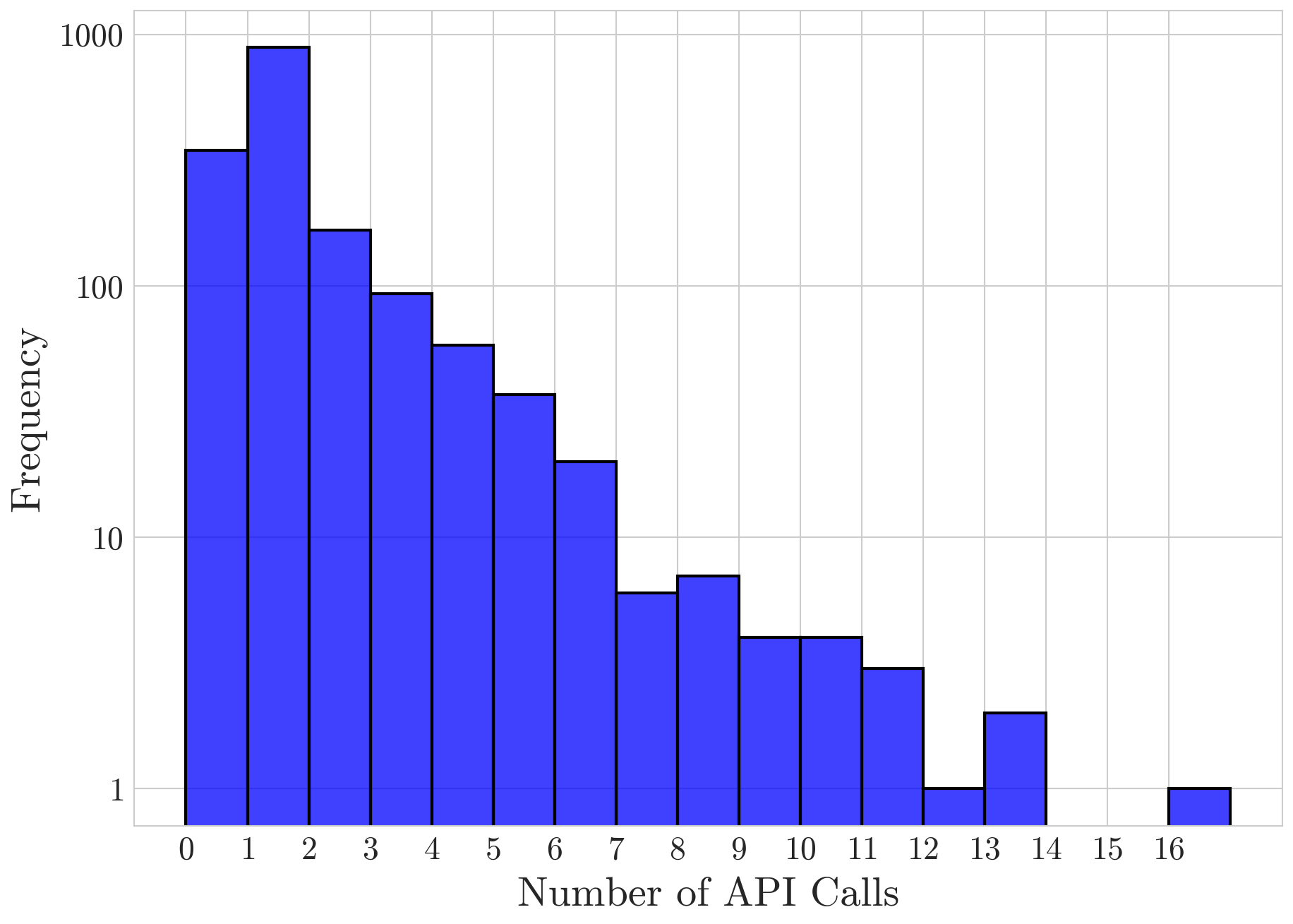}
    \caption{Distribution of API calls per query on OpenEQA}
    \label{fig:api_call_dist}
    \vspace{-0.4cm}
\end{figure}

Analysis of GraphPad's API usage (Fig.~\ref{fig:api_call_dist}) shows that 95\% of questions are answered with 5 or fewer API calls, with an average of 1.9 calls per question. The distribution indicates that in most cases, the system requires relatively few updates to answer questions, though we observe a long tail of more complex queries requiring additional refinement steps.

The peak at zero calls represents questions that could be answered using only the initial scene representation without any additional API calls, suggesting that for some question types, the initial structured memory provides sufficient context.

\section{Limitations}
\label{sec:limitations}

While GraphPad demonstrates the value of editable 3D scene representations, several important limitations affect its current implementation:

\begin{itemize}
\item \textbf{Error propagation in detection.} Our current implementation lacks a verification mechanism for object detection quality. When the VLM misidentifies an object or relation, this error becomes part of the scene graph and can negatively affect downstream reasoning. 

\item \textbf{API design constraints.} The three operations (find, analyze objects, analyze frame) represent our initial attempt at a minimal API set, but we have not rigorously evaluated whether this is optimal. Different reasoning tasks might benefit from specialized APIs not explored in this work.

\item \textbf{Computational overhead.} Each API call requires a full VLM inference pass, adding significant latency (typically 2-3 seconds per call in our implementation). This latency currently limits GraphPad's applicability to real-time systems.

\item \textbf{Domain generalization.} Our APIs were designed specifically for question answering about static scenes. We have not tested their effectiveness for other domains like manipulation planning or navigation in dynamic environments.

\item \textbf{Scalability limits.} As scene graphs grow larger, both the prompt size and reasoning complexity increase. Our experiments were limited to medium-sized home environments; performance in larger spaces remains untested.
\end{itemize}

\section{Conclusion}
\label{sec:conclusion}

We presented GraphPad, a system that enables VLMs to update 3D scene graphs during inference through language-callable functions. On OpenEQA, GraphPad improved accuracy by 3.0 percentage points over an image-only baseline using the same VLM while requiring fewer input frames.

Our experiments suggest several insights for 3D language-vision systems. First, static structured representations appear to benefit from targeted, task-specific refinement. Second, language-guided perception may offer a middle ground between exhaustive preprocessing and purely reactive vision. Third, the efficiency of GraphPad (averaging under 2 API calls per question) indicates that targeted scene exploration can be a practical strategy.

For future 3D vision-language-action systems, these results suggest investigating how reasoning agents might direct their own perception in service of task goals. Extending GraphPad's approach to manipulation planning, navigation, and dynamic scenes could help bridge the gap between language understanding and effective action in 3D environments.

{
    \small
    \bibliographystyle{ieeenat_fullname}
    \bibliography{main}

\begin{thebibliography}{30}
\providecommand{\natexlab}[1]{#1}
\providecommand{\url}[1]{\texttt{#1}}
\expandafter\ifx\csname urlstyle\endcsname\relax
  \providecommand{\doi}[1]{doi: #1}\else
  \providecommand{\doi}{doi: \begingroup \urlstyle{rm}\Url}\fi

\bibitem[Anwar et~al.(2024)Anwar, Welsh, Biswas, Pouya, and Chang]{anwar2024remembr}
Abrar Anwar, John Welsh, Joydeep Biswas, Soha Pouya, and Yan Chang.
\newblock Remembr: Building and reasoning over long-horizon spatio-temporal memory for robot navigation.
\newblock \emph{arXiv preprint arXiv:2409.13682}, 2024.

\bibitem[Chiang et~al.(2024)Chiang, Xu, Fu, Jacob, Zhang, Lee, Yu, Schenck, Rendleman, Shah, Xia, Hsu, Hoech, Florence, Kirmani, Singh, Sindhwani, Parada, Finn, Xu, Levine, and Tan]{chiang2024mobilityvlamultimodalinstruction}
Hao-Tien~Lewis Chiang, Zhuo Xu, Zipeng Fu, Mithun~George Jacob, Tingnan Zhang, Tsang-Wei~Edward Lee, Wenhao Yu, Connor Schenck, David Rendleman, Dhruv Shah, Fei Xia, Jasmine Hsu, Jonathan Hoech, Pete Florence, Sean Kirmani, Sumeet Singh, Vikas Sindhwani, Carolina Parada, Chelsea Finn, Peng Xu, Sergey Levine, and Jie Tan.
\newblock Mobility vla: Multimodal instruction navigation with long-context vlms and topological graphs.
\newblock \emph{arXiv preprint arXiv:2407.07775}, 2024.

\bibitem[Dai et~al.(2017)Dai, Chang, Savva, Halber, Funkhouser, and Nie{\ss}ner]{dai2017scannet}
Angela Dai, Angel~X. Chang, Manolis Savva, Maciej Halber, Thomas Funkhouser, and Matthias Nie{\ss}ner.
\newblock Scannet: Richly-annotated 3d reconstructions of indoor scenes.
\newblock In \emph{Proc. Computer Vision and Pattern Recognition (CVPR), IEEE}, 2017.

\bibitem[Das et~al.(2018)Das, Datta, Gkioxari, Lee, Parikh, and Batra]{Das_2018_CVPR}
Abhishek Das, Samyak Datta, Georgia Gkioxari, Stefan Lee, Devi Parikh, and Dhruv Batra.
\newblock Embodied question answering.
\newblock In \emph{Proceedings of the IEEE Conference on Computer Vision and Pattern Recognition (CVPR)}, 2018.

\bibitem[Ding et~al.(2023)Ding, Yang, Xue, Zhang, Bai, and Qi]{ding2023pla}
Runyu Ding, Jihan Yang, Chuhui Xue, Wenqing Zhang, Song Bai, and Xiaojuan Qi.
\newblock Pla: Language-driven open-vocabulary 3d scene understanding.
\newblock In \emph{Proceedings of the IEEE/CVF conference on computer vision and pattern recognition}, pages 7010--7019, 2023.

\bibitem[Greve et~al.(2024)Greve, Büchner, Vödisch, Burgard, and Valada]{greve2023curb}
Elias Greve, Martin Büchner, Niclas Vödisch, Wolfram Burgard, and Abhinav Valada.
\newblock Collaborative dynamic 3d scene graphs for automated driving.
\newblock In \emph{2024 IEEE International Conference on Robotics and Automation (ICRA)}, pages 11118--11124, 2024.

\bibitem[Gu et~al.(2024)Gu, Kuwajerwala, Morin, Jatavallabhula, Sen, Agarwal, Rivera, Paul, Ellis, Chellappa, et~al.]{gu2024conceptgraphs}
Qiao Gu, Ali Kuwajerwala, Sacha Morin, Krishna~Murthy Jatavallabhula, Bipasha Sen, Aditya Agarwal, Corban Rivera, William Paul, Kirsty Ellis, Rama Chellappa, et~al.
\newblock Conceptgraphs: Open-vocabulary 3d scene graphs for perception and planning.
\newblock In \emph{2024 IEEE International Conference on Robotics and Automation (ICRA)}, pages 5021--5028. IEEE, 2024.

\bibitem[Honerkamp et~al.(2024)Honerkamp, B{\"u}chner, Despinoy, Welschehold, and Valada]{honerkamp2024language}
Daniel Honerkamp, Martin B{\"u}chner, Fabien Despinoy, Tim Welschehold, and Abhinav Valada.
\newblock Language-grounded dynamic scene graphs for interactive object search with mobile manipulation.
\newblock \emph{IEEE Robotics and Automation Letters}, 2024.

\bibitem[Huang et~al.(2023)Huang, Mees, Zeng, and Burgard]{huang23vlmaps}
Chenguang Huang, Oier Mees, Andy Zeng, and Wolfram Burgard.
\newblock Visual language maps for robot navigation.
\newblock In \emph{Proceedings of the IEEE International Conference on Robotics and Automation (ICRA)}, London, UK, 2023.

\bibitem[Hughes et~al.(2022)Hughes, Chang, and Carlone]{hughes2022hydra}
Nathan Hughes, Yun Chang, and Luca Carlone.
\newblock Hydra: A real-time spatial perception system for 3d scene graph construction and optimization.
\newblock \emph{arXiv preprint arXiv:2201.13360}, 2022.

\bibitem[Li et~al.(2024)Li, Yu, She, Yu, Lan, Zhu, Hu, and Xu]{li2024llm}
Wenhao Li, Zhiyuan Yu, Qijin She, Zhinan Yu, Yuqing Lan, Chenyang Zhu, Ruizhen Hu, and Kai Xu.
\newblock Llm-enhanced scene graph learning for household rearrangement.
\newblock In \emph{SIGGRAPH Asia 2024 Conference Papers}, pages 1--11, 2024.

\bibitem[Liu et~al.(2024)Liu, Guo, Warke, Chintala, Paxton, Shafiullah, and Pinto]{liu2024dynamem}
Peiqi Liu, Zhanqiu Guo, Mohit Warke, Soumith Chintala, Chris Paxton, Nur Muhammad~Mahi Shafiullah, and Lerrel Pinto.
\newblock Dynamem: Online dynamic spatio-semantic memory for open world mobile manipulation.
\newblock \emph{arXiv preprint arXiv:2411.04999}, 2024.

\bibitem[Maggio et~al.(2024)Maggio, Chang, Hughes, Trang, Griffith, Dougherty, Cristofalo, Schmid, and Carlone]{maggio2024clio}
Dominic Maggio, Yun Chang, Nathan Hughes, Matthew Trang, Dan Griffith, Carlyn Dougherty, Eric Cristofalo, Lukas Schmid, and Luca Carlone.
\newblock Clio: Real-time task-driven open-set 3d scene graphs.
\newblock \emph{arXiv preprint arXiv:2404.13696}, 2024.

\bibitem[Majumdar et~al.(2024)Majumdar, Ajay, Zhang, Putta, Yenamandra, Henaff, Silwal, Mcvay, Maksymets, Arnaud, et~al.]{majumdar2024openeqa}
Arjun Majumdar, Anurag Ajay, Xiaohan Zhang, Pranav Putta, Sriram Yenamandra, Mikael Henaff, Sneha Silwal, Paul Mcvay, Oleksandr Maksymets, Sergio Arnaud, et~al.
\newblock Openeqa: Embodied question answering in the era of foundation models.
\newblock In \emph{Proceedings of the IEEE/CVF Conference on Computer Vision and Pattern Recognition}, pages 16488--16498, 2024.

\bibitem[Peng et~al.(2023)Peng, Genova, Jiang, Tagliasacchi, Pollefeys, Funkhouser, et~al.]{peng2023openscene}
Songyou Peng, Kyle Genova, Chiyu Jiang, Andrea Tagliasacchi, Marc Pollefeys, Thomas Funkhouser, et~al.
\newblock Openscene: 3d scene understanding with open vocabularies.
\newblock In \emph{Proceedings of the IEEE/CVF conference on computer vision and pattern recognition}, pages 815--824, 2023.

\bibitem[Pichai et~al.(2024)Pichai, Hassabis, and Kavukcuoglu]{pichai2024gemini}
Sundar Pichai, Demis Hassabis, and Koray Kavukcuoglu.
\newblock Introducing {Gemini} 2.0: our new {AI} model for the agentic era.
\newblock \url{https://blog.google/technology/ai/introducing-gemini-2/}, 2024.
\newblock Google Blog, \url{https://blog.google/technology/ai/introducing-gemini-2/}.

\bibitem[Rajvanshi et~al.(2024)Rajvanshi, Sikka, Lin, Lee, pang Chiu, and Velasquez]{rajvanshi2024saynav}
Abhinav Rajvanshi, Karan Sikka, Xiao Lin, Bhoram Lee, Han pang Chiu, and Alvaro Velasquez.
\newblock Saynav: Grounding large language models for dynamic planning to navigation in new environments.
\newblock In \emph{34th International Conference on Automated Planning and Scheduling}, 2024.

\bibitem[Ramakrishnan et~al.(2021)Ramakrishnan, Gokaslan, Wijmans, Maksymets, Clegg, Turner, Undersander, Galuba, Westbury, Chang, Savva, Zhao, and Batra]{ramakrishnan2021hm3d}
Santhosh~Kumar Ramakrishnan, Aaron Gokaslan, Erik Wijmans, Oleksandr Maksymets, Alexander Clegg, John~M Turner, Eric Undersander, Wojciech Galuba, Andrew Westbury, Angel~X Chang, Manolis Savva, Yili Zhao, and Dhruv Batra.
\newblock Habitat-matterport 3d dataset ({HM}3d): 1000 large-scale 3d environments for embodied {AI}.
\newblock In \emph{Thirty-fifth Conference on Neural Information Processing Systems Datasets and Benchmarks Track}, 2021.

\bibitem[Rana et~al.(2023)Rana, Haviland, Garg, Abou-Chakra, Reid, and Suenderhauf]{rana2023sayplan}
Krishan Rana, Jesse Haviland, Sourav Garg, Jad Abou-Chakra, Ian Reid, and Niko Suenderhauf.
\newblock Sayplan: Grounding large language models using 3d scene graphs for scalable task planning.
\newblock In \emph{7th Annual Conference on Robot Learning}, 2023.

\bibitem[Rosinol et~al.(2021)Rosinol, Violette, Abate, Hughes, Chang, Shi, Gupta, and Carlone]{kimera}
A. Rosinol, A. Violette, M. Abate, N. Hughes, Y. Chang, J. Shi, A. Gupta, and L. Carlone.
\newblock {K}imera: from {SLAM} to spatial perception with {3D} dynamic scene graphs.
\newblock In \emph{arXiv preprint arXiv:2101.06894}, 2021.
\newblock \url{https://arxiv.org/pdf/2101.06894.pdf}.

\bibitem[Saxena et~al.(2024)Saxena, Buchanan, Paxton, Chen, Vaskevicius, Palmieri, Francis, and Kroemer]{saxena2024grapheqa}
Saumya Saxena, Blake Buchanan, Chris Paxton, Bingqing Chen, Narunas Vaskevicius, Luigi Palmieri, Jonathan Francis, and Oliver Kroemer.
\newblock Grapheqa: Using 3d semantic scene graphs for real-time embodied question answering.
\newblock \emph{arXiv preprint arXiv:2412.14480}, 2024.

\bibitem[Shah et~al.(2024)Shah, Yu, Zhu, Zhu, and Martín-Martín]{shah2024bumbleunifyingreasoningacting}
Rutav Shah, Albert Yu, Yifeng Zhu, Yuke Zhu, and Roberto Martín-Martín.
\newblock Bumble: Unifying reasoning and acting with vision-language models for building-wide mobile manipulation.
\newblock \emph{arXiv preprint arXiv:2410.06237}, 2024.

\bibitem[Takmaz et~al.(2024)Takmaz, Delitzas, Sumner, Engelmann, Wald, and Tombari]{takmaz2024search3d}
Ayca Takmaz, Alexandros Delitzas, Robert~W Sumner, Francis Engelmann, Johanna Wald, and Federico Tombari.
\newblock Search3d: Hierarchical open-vocabulary 3d segmentation.
\newblock \emph{arXiv preprint arXiv:2409.18431}, 2024.

\bibitem[Tang et~al.(2025)Tang, Wang, Deng, Zheng, Deng, and Yue]{tang2025openinopenvocabularyinstanceorientednavigation}
Yujie Tang, Meiling Wang, Yinan Deng, Zibo Zheng, Jingchuan Deng, and Yufeng Yue.
\newblock Openin: Open-vocabulary instance-oriented navigation in dynamic domestic environments.
\newblock \emph{arXiv preprint arXiv:2501.04279}, 2025.

\bibitem[Wang et~al.(2024)Wang, Yu, Zhao, Sun, Hou, Liang, Hu, Han, and Gan]{wang2024karma}
Zixuan Wang, Bo Yu, Junzhe Zhao, Wenhao Sun, Sai Hou, Shuai Liang, Xing Hu, Yinhe Han, and Yiming Gan.
\newblock Karma: Augmenting embodied ai agents with long-and-short term memory systems.
\newblock \emph{arXiv preprint arXiv:2409.14908}, 2024.

\bibitem[Werby et~al.(2024)Werby, Huang, B{\"u}chner, Valada, and Burgard]{werby2024hierarchical}
Abdelrhman Werby, Chenguang Huang, Martin B{\"u}chner, Abhinav Valada, and Wolfram Burgard.
\newblock Hierarchical open-vocabulary 3d scene graphs for language-grounded robot navigation.
\newblock In \emph{First Workshop on Vision-Language Models for Navigation and Manipulation at ICRA 2024}, 2024.

\bibitem[Xiao et~al.(2024)Xiao, Liu, Zhang, Muennighoff, Lian, and Nie]{xiao2024cpackpackedresourcesgeneral}
Shitao Xiao, Zheng Liu, Peitian Zhang, Niklas Muennighoff, Defu Lian, and Jian-Yun Nie.
\newblock C-pack: Packed resources for general chinese embeddings, 2024.

\bibitem[Xie et~al.(2024)Xie, Min, Zhang, Xu, Bajaj, Salakhutdinov, Johnson-Roberson, and Bisk]{xie2024embodied}
Quanting Xie, So~Yeon Min, Tianyi Zhang, Kedi Xu, Aarav Bajaj, Ruslan Salakhutdinov, Matthew Johnson-Roberson, and Yonatan Bisk.
\newblock Embodied-rag: General non-parametric embodied memory for retrieval and generation.
\newblock \emph{arXiv preprint arXiv:2409.18313}, 2024.

\bibitem[Yang et~al.(2024)Yang, Yang, Zhou, Chen, Zhang, Du, and Gan]{yang20243dmem3dscenememory}
Yuncong Yang, Han Yang, Jiachen Zhou, Peihao Chen, Hongxin Zhang, Yilun Du, and Chuang Gan.
\newblock 3d-mem: 3d scene memory for embodied exploration and reasoning.
\newblock \emph{arXiv preprint arXiv:2411.17735}, 2024.

\bibitem[Zhang et~al.(2024)Zhang, Qu, Patil, Cadena, and Hutter]{zhang2024tag}
Mike Zhang, Kaixian Qu, Vaishakh Patil, Cesar Cadena, and Marco Hutter.
\newblock Tag map: A text-based map for spatial reasoning and navigation with large language models.
\newblock \emph{arXiv preprint arXiv:2409.15451}, 2024.

\end{thebibliography}
}


\end{document}